\documentclass[runningheads]{llncs}
\usepackage{graphicx}
\usepackage{amsmath,amssymb} 
\usepackage{color}
\usepackage[width=122mm,left=12mm,paperwidth=146mm,height=193mm,top=12mm,paperheight=217mm]{geometry}

\usepackage{algorithm}
\usepackage{algpseudocode}

\usepackage{amssymb}
\usepackage{pifont}

\newcommand{\cmark}{\ding{51}}%
\newcommand{\xmark}{\ding{55}}%

\usepackage{epsfig}
\usepackage{graphicx}
\usepackage{comment}

\begin{document}
\pagestyle{headings}
\mainmatter
\def\ECCV18SubNumber{}  

\title{An Integral Pose Regression System for the ECCV2018 PoseTrack Challenge} 

\titlerunning{Integral Pose Regression System}

\authorrunning{X. Sun, C. Li and S. Lin}


\author{Xiao Sun\inst{1} \and
Chuankang Li\inst{2} \and
Stephen Lin\inst{1}}

\institute{
Microsoft Research\\
\email{\{xias, stevelin\}@microsoft.com}\\ 
\and
Zhejiang University\\
\email{lck-cad@zju.edu.cn}}

\maketitle

\begin{abstract} For the ECCV 2018 PoseTrack Challenge, we present a 3D human pose estimation system based mainly on the integral human pose regression method. We show a comprehensive ablation study to examine the key performance factors of the proposed system. Our system obtains \emph{47mm} MPJPE on the CHALL\_H80K \emph{test} dataset, placing second in the ECCV2018 3D human pose estimation challenge. Code will be released to facilitate future work.

\end{abstract}



\section{Introduction}
\label{sec.introduction}

The \emph{ECCV2018 3D Human Pose Estimation Challenge} evaluates proposed methods for estimating 3D key points of people from monocular RGB images. The challenge is based on the CHALL\_H80K subset of the popular \emph{Human3.6M~\cite{ionescu2011latent,ionescu2014human3}} benchmark. CHALL\_H80K contains 80K 3D human poses and corresponding images from 10 professional actors (6 male, 4 female) and 15 scenarios (discussion, smoking, taking photo, talking on the phone, etc.). Among them, 5 subjects (36K frames) are used for training, 2 subjects (20K frames) for validation and 4 subjects (24K frames) for testing. The images are captured in a controlled environment in which the subjects and background have a simple appearance. In CHALL\_H80K, only 3D pose ground-truth is provided, using the root (pelvis) joint as the origin, expressed in millimeters (mm). For evaluation, the mean per joint position error (\emph{MPJPE})~\cite{ionescu2014human3} metric is used.




Human pose estimation has been extensively studied~\cite{ionescu2014human3,andriluka20142d,lin2014microsoft}. Recent years have seen significant progress on the problem due to advances in deep convolutional neural networks (CNNs). The best performing methods on 2D pose estimation are all detection-based~\cite{mpiiwebpage}. They generate a likelihood heat map for each joint and locate the joint as the point with the maximum likelihood in the map. Promising extensions of the heat map approach have been presented for 3D pose estimation~\cite{pavlakos2016coarse}. Most recently, Sun et al.~\cite{sun2017integral} replaced the argmax post-processing with an integral formulation to tackle the quantization problem and enable end-to-end learning. This approach currently achieves the highest 3D pose estimation performance.

For the ECCV 2018 PoseTrack Challenge, we present a 3D human pose estimation system based mainly on the \emph{integral human pose regression}~\cite{sun2017integral} method. Within the rules of the challenge, some external data is used in our system besides CHALL\_H80K. Specifically, 2D human pose data from MPII~\cite{andriluka20142d} and COCO~\cite{lin2014microsoft} are used for both camera model estimation and 2D/3D mixed data training. COCO object bounding box data is used to train a person box detector for a two-stage top-down pose estimation paradigm similar to~\cite{sun2017integral,papandreou2017towards}. ImageNet~\cite{deng2009imagenet} classification data is employed for pre-training the backbone networks. No other external data is used. Our system obtains \emph{47mm} MPJPE on the CHALL\_H80K \emph{test} dataset, placing second in the ECCV2018 3D human pose estimation challenge~\cite{hm36webpage}. Code\footnote{https://github.com/JimmySuen/integral-human-pose} will be released to facilitate future work.

\section{Overview}
\label{sec.overview}

\begin{figure*}
\centering
\includegraphics [width=1.0\linewidth] {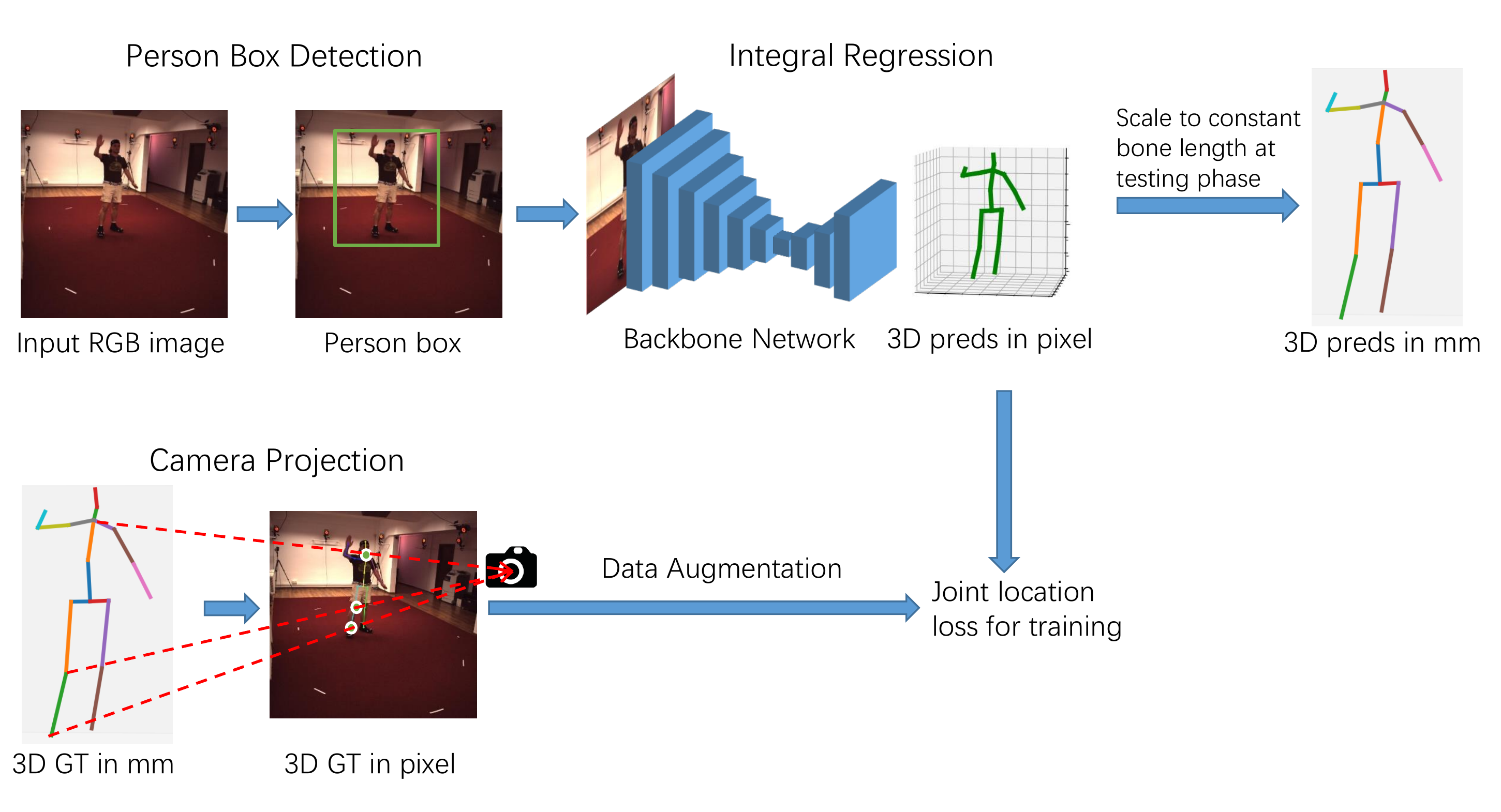}
\caption{Overview of 3D pose estimation framework.}
\label{fig.overview}
\end{figure*}

Figure~\ref{fig.overview} provides an overview of our system. It contains three components. First, a \emph{person box detection} component roughly localizes the person in the input RGB image. Second, a \emph{camera projection} component is used to project 3D ground truth to the image coordinate system, as done in per-pixel/voxel classification based learning methods. These two components are used for data pre-processing. Third, the core CNN based learning component applies \emph{Integral Regression}~\cite{sun2017integral} to perform 3D human pose estimation.

\paragraph{\textbf{Person Box Detection}} Instead of predicting the 3D key points from the original image directly, we follow a two-stage top-down paradigm similar to~\cite{sun2017integral,papandreou2017towards}. First, a person box detector is used to roughly localize the person. Then, a normalized local image patch is generated by cropping the original image with the box and resizing it to a fixed image resolution. The normalized local image patches are used as the final inputs of the CNN model. Since much of the background area is detected and removed in this first stage, the accuracy of the following key point detection model can be greatly improved. As shown in Table~\ref{table.ablation}, we obtain a $25.4\%$ relative improvement (MPJPE decreases from $115.9 mm$ to $86.5 mm$) by using the person box detector.

\paragraph{\textbf{Camera Projection Model}} 
Following the detection based pose estimation paradigm, which is essentially a per pixel/voxel classification task, we first project the ground truth to the image coordinate system (in pixels). Since the CHALL\_H80K dataset provides the 3D ground truth only in mm, we assume a \emph{weak perspective} camera projection model and then estimate the camera model parameters by matching the 2D projection of the 3D ground truth with a coarse 2D key points estimation result, which is obtained from a pre-trained 2D model using external 2D data (MPII). 

At the testing phase, the camera model parameters are unknown. Hence, we are not able to recover the final 3D key points using camera back-projection. Instead, we scale the predicted 3D key points in pixels to conform to a particular average bone length to obtain the final prediction in mm. Different choices of bone length significantly affect the final performance, as shown in Table~\ref{table.bone_length}. 

\begin{table}
\caption{Effect of using different bone length at testing phase.}
\begin{center}
\begin{tabular}{l | l | l | l | l}
\hline
Bone Length Type& Per-frame & Avg Val 	& Avg Train & Avg Train+Val \\
MPJPE(mm) 	& \textbf{60.0} &  64.6	& 65.6	& 65.2 \\
\hline
\end{tabular}
\end{center}
\label{table.bone_length}
\end{table}

Not surprisingly, using the \emph{per-frame} ground truth average bone length substantially outperforms using other average bone lengths, such as those computed from the validation or training sets. In real applications, it is possible for users to provide this personalized information to strengthen the system. In this challenge, however, bone length information is not given on the \emph{test} dataset, which leaves us with using an average bone length estimated from a given dataset. In practice, we use \emph{Avg Train+Val} bone length in all of our experiments.

\paragraph{\textbf{Integral Regression}} Sun et al.~\cite{sun2017integral} recently presented a simple and effective integral regression method that unifies the heat map representation and joint regression approaches in a manner that preserves the merits of both. It replaces the non-differentiable argmax post-processing with the differentiable integral operation, thus allowing end-to-end training and producing continuous output to solve the quantization problem. Moreover, they generalize this approach to the 3D human pose estimation problem for the first time and achieve the state of the art result on the Human3.6M dataset.

We use the Integral Regression method to train our 3D human pose estimation model. In~\cite{sun2017integral}, several variants of Integral Regression are proposed and investigated according to different heat map loss types and joint loss types. In their experiments, L1 joint loss without heat map loss pre-training achieves the best performance for the 3D pose estimation problem. We adopt this variant in all of our experiments. 

The 3D pose estimation performance can be significantly improved by adding abundant 2D pose data to the training \cite{sun2017integral,sun2017compositional}. This is feasible because the integral formulation generates $x,y,z$ predictions individually and maintains differentiability. In addition, our camera model estimation component projects 3D ground truth to the image coordinate system. As shown in Table~\ref{table.ablation}, we get $28.1\%$ relative improvement (MPJPE decreases from $86.5 mm$ to $62.2 mm$) by using external MPII 2D pose data for mixed 2D/3D training.

\section{Experiments}
\label{sec.imp}

\paragraph{\textbf{Training}} ResNet~\cite{he2016deep} is adopted as the backbone network and is pre-trained on the ImageNet classification dataset~\cite{deng2009imagenet}. A normal distribution with 1e-3 standard deviation is used to initialize the head network parameters as in \cite{sun2017integral} for integral regression. PyTorch~\cite{paszke2017automatic} is used for implementation. Adam is employed for optimization. Data augmentation includes random translation ($\pm2\%$ of the image size), scale ($\pm25\%$), rotation ($\pm30$ degrees) and flip. In all experiments, the base learning rate is 1e-4. It drops to 1e-6 when the loss on the validation set saturates. The training iterations proceed until performance on the validation set saturates. Four GPUs are utilized. The mini-batch size is 64, and batch-normalization~\cite{ioffe2015batch} is used. For our ablation study, the CHALL\_H80K \emph{train} and \emph{val} datasets are used for training and evaluation, respectively. For the final challenge result on the CHALL\_H80K \emph{test} dataset, both \emph{train} and \emph{val} datasets are used for training. Other training details are provided with the individual experiments.

\paragraph{\textbf{Ablation Study}} Table~\ref{table.ablation} shows a comprehensive ablation study to examine the key performance factors of the proposed framework, including the two-stage paradigm using person box detection (25.4\% relative improvement), and 2D and 3D mixed data training (28.1\% relative improvement). Additionally, the effect of different training strategies including deeper backbone networks (1.6\% relative improvement) and larger image resolution (4.4\% relative improvement), together with testing strategies including flip testing (1.0\% relative improvement) and model ensemble (2.8\% relative improvement) are investigated.

\begin{table}
\begin{minipage}{1.0\textwidth}
\centering
\caption{Ablation study. All models are trained on CHALL\_H80K \emph{train} dataset and evaluated on CHALL\_H80K \emph{val} dataset.}
\resizebox{!}{1.55cm}{
\begin{tabular}{l | l | l | l | l | l | l | l }
\hline
 description & box det. & dataset & backbone & patch size & flip test & ensemble & MPJPE(mm) \\
original baseline & \xmark  & HM36   	& ResNet-50 &  - 		& \xmark & \xmark & $115.9$\\
+person box det. & \cmark  & HM36 & ResNet-50    	&  256*256 & \xmark & \xmark & $86.5_{\downarrow25.4\%}$\\
+MPII data & \cmark  & HM36+MPII  & ResNet-50 &  256*256 & \xmark & \xmark & $62.2_{\downarrow28.1\%}$\\
+deeper & \cmark & HM36+MPII & ResNet-152  &  256*256 & \xmark & \xmark & $61.2_{\downarrow1.6\%}$\\
+larger image & \cmark & HM36+MPII & ResNet-152  &  288*384 & \xmark & \xmark & $58.5_{\downarrow4.4\%}$\\
+COCO data & \cmark & HM36+MPII+COCO & ResNet-152  &  288*384 & \xmark & \xmark & $57.5_{\downarrow1.7\%}$\\
+flip test & \cmark & HM36+MPII+COCO & ResNet-152  & 288*384 & \cmark & \xmark & $56.9_{\downarrow1.0\%}$\\
+model ensemble& \cmark & HM36+MPII+COCO & ResNet-152 & 288*384 & \cmark & \cmark & $55.3_{\downarrow2.8\%}$\\
\hline
\end{tabular}}
\label{table.ablation}
\end{minipage}
\end{table}

\paragraph{\textbf{Challenge Result}} We use the best setting in Table~\ref{table.ablation}, namely the last entry, to produce our final result on the CHALL\_H80K \emph{test} dataset. Both \emph{train} and \emph{val} datasets are used for training. Our system obtains \emph{47mm} MPJPE, giving it second place in the ECCV2018 3D human pose estimation challenge~\cite{hm36webpage}.

\bibliographystyle{splncs}
\bibliography{egbib}
\end{document}